%% file: 0RAL_8page.tex
\documentclass[letterpaper, 10 pt, conference]{ieeeconf}  
\IEEEoverridecommandlockouts                              
\usepackage{float} 
\usepackage{graphicx}
\usepackage{subcaption}
\usepackage{booktabs}
\usepackage{makecell}
\usepackage{siunitx, mhchem}
\usepackage{algorithm}
\usepackage{algpseudocode}
\usepackage{amsmath}
\usepackage{float}
\usepackage{epstopdf}
\usepackage{color}
\usepackage{hyperref}
\usepackage[capitalise]{cleveref}
\usepackage{cite}
\usepackage[colorinlistoftodos]{todonotes}
\usepackage{amsfonts}

\title{\textbf{Socially-Compatible Behavior Design of Autonomous Vehicles \\ with Verification on Real Human Data}}

\author{Letian Wang$^{1,*}$, Liting Sun$^{2,*}$, Masayoshi Tomizuka$^2$, and Wei Zhan$^2$
\thanks{$^*$The authors are equally contributed.}
\thanks{$^{1}$Letian Wang is with the Department of Dynamics and Control, Beihang University, Beijing, China. The work was conducted during his visit at University of California, Berkeley.}
\thanks{$^{2}$L. Sun, M. Tomizuka and W. Zhan are with the Department of Mechanical Engineering, University of California, Berkeley, Berkeley, California, U.S.A. Corresponding email: {\tt\small litingsun@berkeley.edu}}}
\begin{document}

\maketitle

\begin{abstract}
As more and more autonomous vehicles (AVs) are being deployed on public roads, designing socially compatible behaviors for them is becoming increasingly important. In order to generate safe and efficient actions, AVs need to not only predict the future behaviors of other traffic participants, but also be aware of the uncertainties associated with such behavior prediction. In this paper, we propose an uncertain-aware integrated prediction and planning (UAPP) framework. It allows the AVs to infer the characteristics of other road users online and generate behaviors optimizing not only their own rewards, but also their courtesy to others, and their confidence regarding the prediction uncertainties. We first propose the definitions for courtesy and confidence. Based on that, their influences on the behaviors of AVs in interactive driving scenarios are explored. Moreover, we evaluate the proposed algorithm on naturalistic human driving data by comparing the generated behavior against ground truth. Results show that the online inference can significantly improve the human-likeness of the generated behaviors. Furthermore, we find that human drivers show great courtesy to others, even for those without right-of-way. We also find that such driving preferences vary significantly in different cultures.

\end{abstract}

\input{1introduction.tex}

\input{2problem_formulation.tex}
\input{3reward_design.tex}
\input{4estimate.tex}
\input{5experiments.tex}
\section{Discussion}\label{sec:conclusion}
{\bf{Summary.}} In this paper, we proposed an integrated prediction and planning framework considering behavior uncertainties. We quantified two social factors, i.e., courtesy and confidence as reward features, and integrated them with the self-interest of autonomous vehicles to generate socially compatible behaviors. Via extensive simulations and systematic analysis, we found that such social factors can encourage the generation of different driving styles. We also proposed an online reward estimation algorithm to infer an individual's preference over the three social factors. Verification on real human data showed that the proposed estimation algorithm and the planning framework can not only help us gain more knowledge about humans' driving behaviors in different traffic conditions and cultures, but also can help generate more human-like behaviors.

\textbf{Future Work.} We will further extend the work in multiple directions, such as 1) extending the algorithm to multi-agent scenarios and relaxing assumptions in game theory, 2) exploring the conditions of policy switch, and 3) comparing the algorithm with other deep learning approaches.

\section*{ACKNOWLEDGEMENT}
We thank Wilko Shwarting and Qingyun Wang for insightful discussions.
\bibliographystyle{IEEEtran.bst}
\bibliography{ref.bib}  
\end{document}

%% file: 1introduction.tex
\section{Introduction}
More and more autonomous vehicles (AVs) are being deployed on public roads, sharing the space with other traffic participants. In order to safely and efficiently interact with other entities (e.g, human drivers and pedestrians), AVs need to generate behaviors that are not only optimal in terms of their own interests, but also showing respect to others in a human-like way \cite{wang2020social}. Such socially-compatible behaviors are of critical importance for the safe deployment of AVs in mixed autonomy. For instance, cautious but unnecessary stops at intersections might cause rear-end accidents \cite{zhan2016nonconservative}, and aggressive ramp merging might lead to collisions. 

To design socially-compatible behaviors, the first two questions to answer are 1) what social factors should be considered, and 2) how we should quantify them and integrate them into the behavior generation framework of AVs. There have been many works in the domain of human-robot interaction (HRI) trying to tackle one or both questions. In terms of methodology, we categorize them into three groups. The first group explicitly imitates some well-established social behaviors of humans via imitation learning or model-based behavior generation \cite{ferrer2014proactive, ratsamee2013human, sun2018fast, sun2019behavior}. One popular example of model-based approaches is to generate robot behaviors via the social force model \cite{helbing1995social} as in \cite{ferrer2014proactive}\cite{ratsamee2013human}. Another example is the social perception behavior in \cite{sun2019behavior} which imitates humans' capability of extracting useful information from others' actions. The second group, on the other hand, quantifies social factors as reward features for the robots and uses inverse reinforcement learning (IRL) to learn the appropriate weights of such features from real human data \cite{sadigh2016information, sun2018courteous, sun2018probabilistic, schwarting2019social, li2020interaction}. Some social factors that have been quantified include active information gathering \cite{sadigh2016information}, deterministic courtesy \cite{sun2018courteous}, and selfishness or altruism via Social Value Orientation \cite{schwarting2019social}. Finally, the third group integrates social factors with deep learning to encourage socially-compliant behaviors\cite{Alahi_2016_CVPR, gupta2018social, jaques2019social, vemula2018social}. Typical examples include Social LSTM \cite{Alahi_2016_CVPR}, Social GAN \cite{gupta2018social}, and social attention\cite{vemula2018social}.
\begin{figure}[t]
	\centering
	\includegraphics[width=0.48\textwidth]{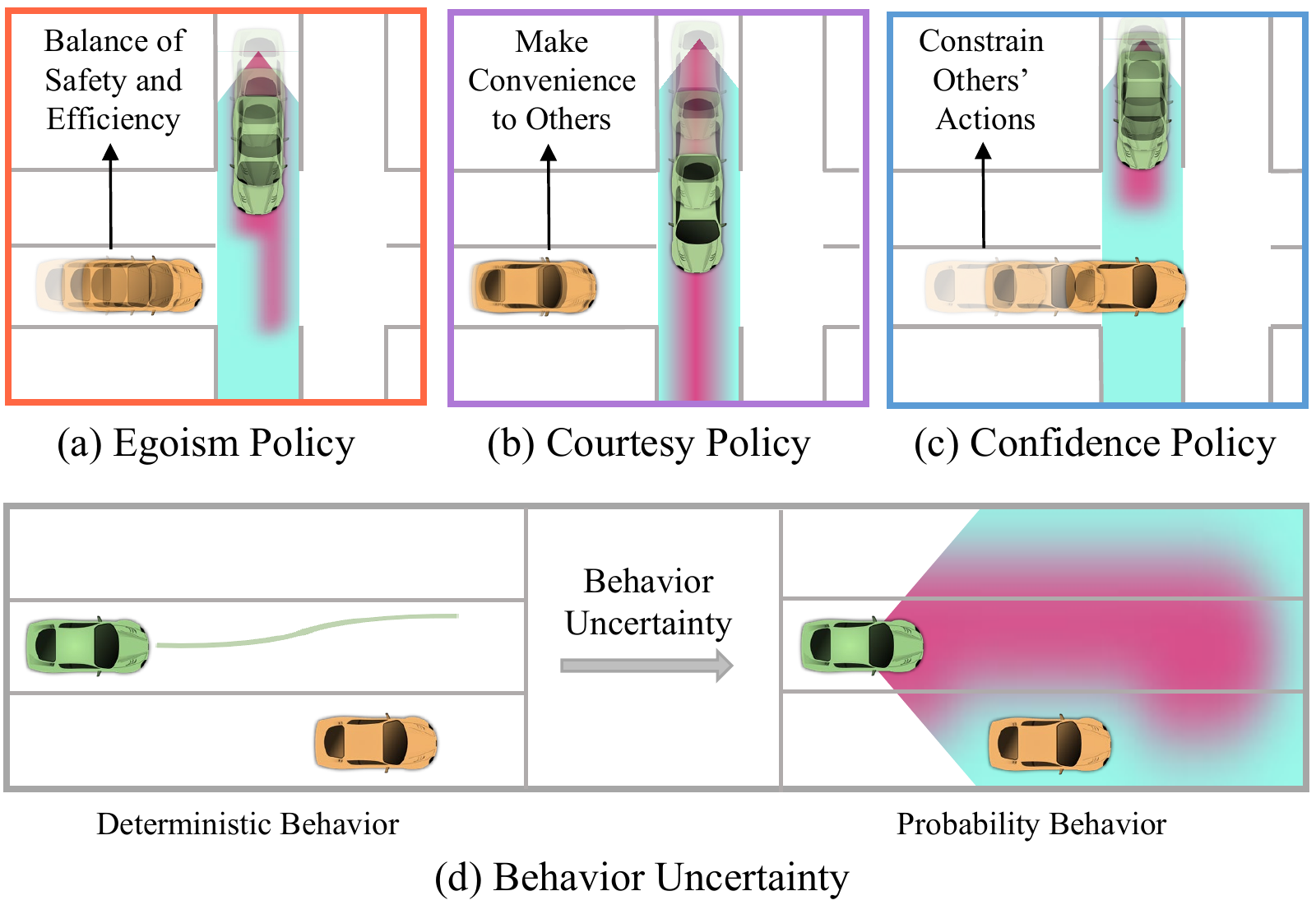}
	\caption{Driving behaviors vary via social policies. (a) Egoistic drivers care about their only utilities. (b) Courteous drivers try to avoid impacting others. (c) Confident drivers prefer to generate interactive results with a highly concentrated distribution rather than a random one.
	}
	\label{fig:motivation}
	\vspace{-2em}
\end{figure}

Our work belongs to the second group for its good interpretability and the ability to learn from data. In particular, we focus on how robot cars should balance their own interests and their impacts on others in the presence of inevitable behavior uncertainties. Note that behaviors of humans and robots are mutually influenced as a closed-loop system. Hence, when behavior uncertainty occurs, the closed-loop dynamics becomes more complicated. First, the distribution of such behavior uncertainty is not fixed but can be changed with different actions of the robots. Second, beyond self-interests and others', the robots (AVs) should also take into consideration the confidence towards uncertain predictions and the corresponding responsive actions in their policies. 

Hence, in this work, we propose an uncertain-aware integrated prediction and planning (UAPP) framework to generate socially-compatible behaviors for autonomous vehicles. \emph{Our key insight is that when behavior uncertainty exists, AVs should optimize for behaviors that consider not only self-expected rewards, but also the impact on the action distributions of others and the self-confidence of such consequences\footnote[1]{Note that many other factors can influence the behavior generation of AVs. For instance, other social factors such as trustworthiness and uncertainties from other functional modules such as perception, localization, and control also impact the behaviors. A comprehensive analysis of all these factors is beyond the scope of this work. We focus on developing an algorithm to quantify some of the social factors emerging from interactions. Hence, we assume perfect conditions for other functional modules.}. Moreover, the preferences over the three aspects should be updated online for different individuals.} 

Our contributions can be summarized as follows: 1) establish an uncertainty-aware integrated prediction and planning (UAPP) framework;
2) formulate the courtesy and confidence factors as rewards in the UAPP framework, and provide a systematic analysis of their influences on AVs' behaviors with varying initial relative relationships between the AVs and other cars, as illustrated in \cref{fig:motivation}; 3) develop an online algorithm to estimate the reward weights of each\title{\big \bf Socially-Compatible Behavior Design of Autonomous Vehicles \\with Verification on Real Human Data} human individual and verify it on real data; and 4) provide inspiring findings on the statistics of how humans drive, such as how often humans switch their preferences among the three policies during one interaction, and how the dominance of driving policy varies with different right-of-ways, traffic rules, and cultures.

%% file: 2problem_formulation.tex
\section{Problem Formulation}
\label{sec:citations}

We consider a two-agent interaction system to investigate the socially-compatible behavior. The two agents include an ego car (denoted by ($\cdot)_{E}$) and an obstacle car (denoted by ($\cdot)_{O}$). They have different right-of-ways.
Let $x_{i}$ and $u_{i}$ denote, respectively, the state and control input of the ego car $(i{=}E)$ and the interacting car $(i{=}O)$. ${\bf{x}}=(x_{E}^T, x_{O}^T)^T$ represents the states of the interaction system, satisfying an
overall dynamics given by:
\begin{equation}
{\bf{x}}^{t+1} = f({\bf{x}}^t, u^t_{E}, u^t_{O}). \label{eq:1}
\end{equation}
\begin{figure}[t!]
	\centering
	\includegraphics[width=0.48\textwidth]{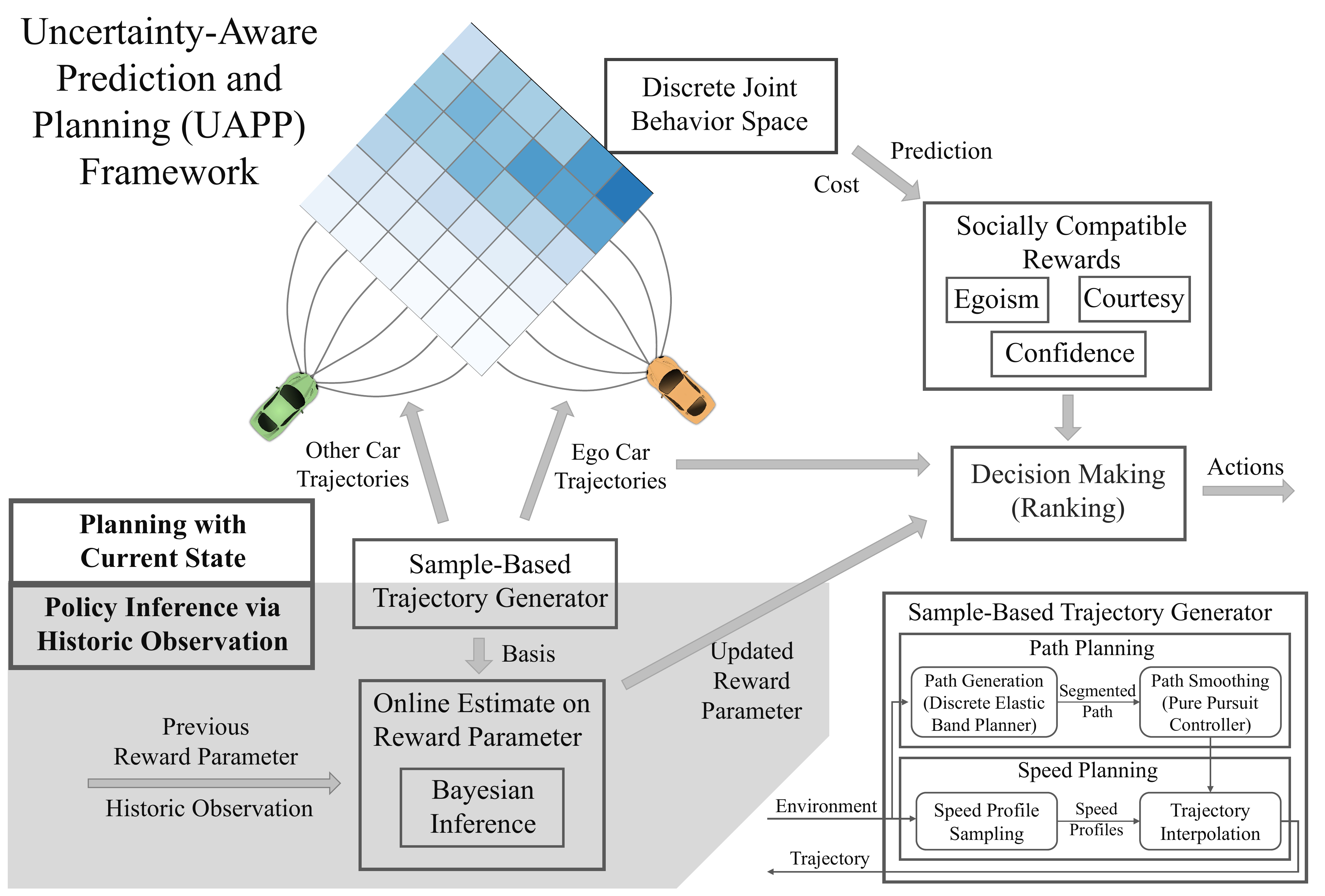}
	\caption{The proposed uncertainty-aware integrated prediction and planning (UAPP) framework consists of several parts: a spatial-temporal sampler to construct discrete joint behavior, a socially compatible reward, and an online algorithm for estimating reward parameter $\mathbf{\lambda}_E$.
	}
	\label{fig:planner}
	\vspace{-2em}
\end{figure}
We assume that both vehicles are noisy optimizers, i.e., their control inputs are optimizing their cumulative reward functions $R_E$ and $R_O$ over the planning horizon. We further assume that the two cars are running a Stackelberg game where the ego car is the leader and the other car is a naive follower, as in \cite{sadigh2016information}\cite{ sun2018courteous}. Let $N$ be the planning horizon. Based on such assumptions and the Model Predictive Control (MPC) framework, at each time step, in the ego car's view, the other vehicle generates its optimal action sequence ${\bf{u}}_O{=}(u_O^0, u_O^1, ..., u_O^{N-1})^T$ via
\begin{equation}
\label{eq:other_vehicle_optimizer}
{\bf{u}}_O^* = \arg\max\limits_{{\bf{u}}_O} R_O({\bf{x}}^0, {{\bf{u}}_O}, {\bf{u}}_E),
\end{equation}
where $\textbf{x}^0$ denotes the current state, and ${\bf{u}}_E$ is a candidate action of the ego car. Thus, considering the behavior uncertainty, the distribution of other vehicle's future behavior $\textbf{u}_O$ can be expressed as a function of the ego car's actions based on the Boltzmann-rational model\cite{luce2012individual}:
\begin{equation}
\label{eq:prediction}
{\bf{u}}_O({\bf{x}}^0, {\bf{u}}_E){\sim} P({\bf{u}}_O|{\bf{x}}^0, {\bf{u}}_E)\propto e^{\beta R_O({\bf{x}}^0, {{\bf{u}}_E}, {{\bf{u}}_O})}.
\end{equation}
$\beta$ controls the degree of rationality of the other vehicle. Without loss of generality, we set $\beta=1$ and omit it for brevity.

With \eqref{eq:prediction} and the assumption that the ego car is a leader in the game, the optimal behavior of the ego car ${\bf{u}}^*_E=(u_E^0, u_E^1, ..., u_E^{N-1})^T$ satisfies
\begin{align}
\label{eq::decision}
{\bf{u}}_E^* =& \arg\max\limits_{{\bf{u}}_E}R_E\left({\bf{x}}^0, {{\bf{u}}_E}, {\bf{u}}_O\sim P({\bf{u}}_O|{\bf{x}}^0, {\bf{u}}_E)\right)\\
=&\arg\max\limits_{{\bf{u}}_E}R_E\left({\bf{x}}^0, {{\bf{u}}_E}\right)\nonumber
\end{align}

\begin{figure*}[!ht]
    \centering
      \includegraphics[width=\textwidth]{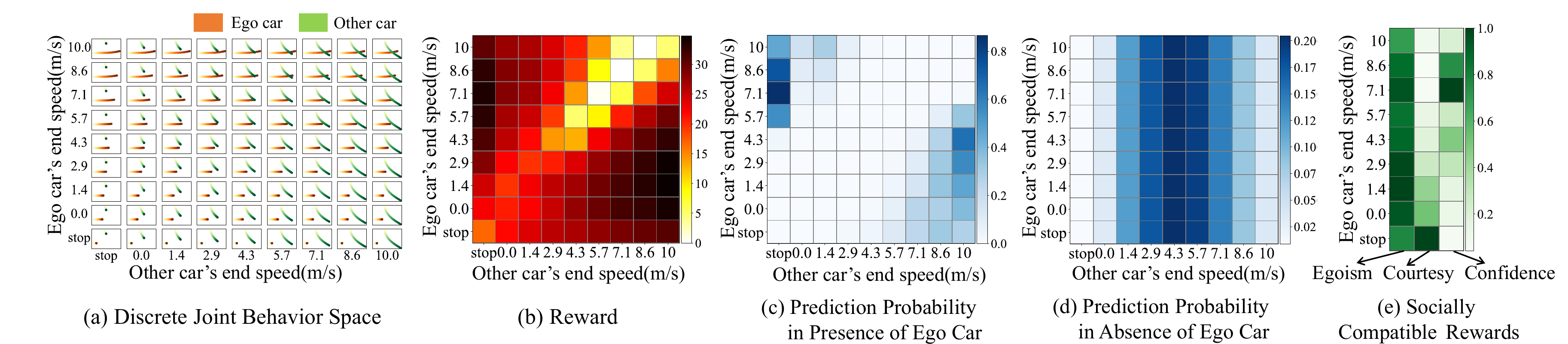}
    \caption{(a) By combining the two car's actions, the agents' joint behavior space is constructed. It captures both discrete decisions and continuous trajectories. (b) The egoism reward of the ego car. (c) and (d) are the probabilities over the joint action spaces with and without the ego car. (e) The socially compatible reward. As shown in (c), the probabilities of the other car's actions are significantly influenced by the ego car's action. 
    }
    \label{fig:reward}
    \vspace{-1.5em}
\end{figure*}
To generate socially compliant behaviors, social factors such as courtesy and confidence should be quantified and formulated as additional features in reward $R_E$. Similar to \cite{levine2012continuous}\cite{sadigh2016planning}, we propose to construct a socially compatible reward for the AVs as a linear combination of three aspects including the regular selfish reward term, the courtesy reward term, and the confidence reward term. Namely, we have
\begin{align}
\label{eq:social_rewards}
	R_E({\bf{x}}^0, {\bf{u}}_E;{\mathbf{\lambda}}_E)&{=}\lambda_{E_{egoism}}R_{E_{egoism}}({\bf{x}}^0, {\bf{u}}_E)\\ 
	&+\lambda_{E_{courtesy}}R_{E_{courtesy}}({\bf{x}}^0, {\bf{u}}_E)\nonumber\\
	&+\lambda_{E_{confidence}}R_{E_{confidence}}({\bf{x}}^0, {\bf{u}}_E)\nonumber
\end{align}
where $R_{E_{egoism}}, R_{E_{courtesy}}, R_{E_{confidence}}$ are, respectively, the ego car's egoism reward, courtesy reward, and confidence reward. Each reward encourages the generation of driving behaviors with different styles. Detailed definitions of them will be introduced in \cref{Sec:reward design}. The reward parameter ${\mathbf{\lambda}}_E{=}(\lambda_{E_{egoism}}, \lambda_{E_{courtesy}}, \lambda_{E_{confidence}})$ captures the trade-off among them.

%% file: 3reward_design.tex
\section{The Integrated Prediction and Planning Framework with Socially Compatible Rewards}
\label{Sec:reward design}
\subsection{The Uncertain-Aware Integrated Prediction and Planning Framework}
The structure of the uncertainty-aware integrated prediction and planning (UAPP) framework is shown in \cref{fig:planner}. It includes a spatial-temporal sampling module to construct the discrete joint behavior space, a socially compatible reward function as defined in \eqref{eq:social_rewards}, and finally an online algorithm for estimating reward parameters $\mathbf{\lambda}_E$. In the discrete joint behavior space, the paired future trajectories of both cars are generated via spatial-temporal sampling \cite{li2020IFAC}\cite{gu2017improved}. One example of such paired trajectories is shown in \cref{fig:reward}(a). \Cref{fig:reward}(b)-(d) demonstrate, respectively, the different reward maps and probabilities for different trajectory samples, and \cref{fig:reward}(e) summarizes the weighted sum of all reward maps.

\subsection{Egoism Reward}
\label{subsec:Cost Reward}
Egoism reward is defined to capture an agent's expected utilities in the presence of behavior uncertainties of others. With the current state $\mathbf{x}^0$ and a pair of candidate actions $(\mathbf{u}_E, \mathbf{u}_O)$, the utility-related reward of an agent is defined as
\begin{equation}
\label{eq:cost function}
    R({\bf{x}}^0, {{\bf{u}}_E}, {{\bf{u}}_O}) =  \theta^T\sum_{t=0}^{N-1}\phi({\bf{x}}^t, u_E^t, u_O^t),
\end{equation}
where $\phi{\in}\mathrm{R}^3$ denotes the three self-interested utilities: efficiency, comfort, and safety\footnote[2]{In this paper, we designed efficiency as keeping the desired speed and staying close to the reference line. Comfort is defined as the longitudinal and lateral acceleration and jerk. Safety is quantified as the relative distance of two cars, and two cars' distance to the intersection point of reference lines.}. $\theta{\in}\mathrm{R}^3$ denotes their relative weights, which can be manually tuned or learned via Inverse Reinforcement Learning\cite{ziebart2008maximum}\cite{levine2012continuous}\cite{wu2020efficient}. 

In order to evaluate the egoism reward, i.e., the expected utilities for the ego vehicle, we have to consider all possible responsive actions $\mathbf{u}_O$ from the other agent. Recalling \eqref{eq:prediction}, we have \begin{equation}
 \label{eq: prediction under ego car} 
 P({\bf{u}}_O|{\bf{x}}^0, {\bf{u}}_E)=\frac{e^{R_O({\bf{u}}_O|{\bf{x}}^0, {\bf{u}}_E)}}{\sum_{{\bf{u}}_O \in {\bf{U}}_O}e^{R_O({\bf{u}}_O|{\bf{x}}^0, {\bf{u}}_E)}},
\end{equation}
where $R_O({\bf{u}}_O|{\bf{x}}^0, {\bf{u}}_E)$ quantifies the other car's rewards as in \eqref{eq:cost function}. Thus, the egoism reward of the ego car's action $\mathbf{u}_E$ is given by:
\begin{equation}\label{eq:egoism_rw}
R_{E_{egoism}}({\bf{x}}^0, {\bf{u}}_E) {=}\mathbb{E}_{{\bf{u}}_O\sim P({\bf{u}}_O|{\bf{x}}^0, {\bf{u}}_E)}\left[R_E({\bf{x}}^0, {\bf{u}}_E, \bf{u}_O)\right].    
\end{equation}

\subsection{Courtesy Reward}
\label{subsec:Courtesy Reward}
There are several ways of modeling courtesy\cite{schwarting2019social}\cite{sun2018courteous}\cite{wang2020social}.
In this paper, we model courtesy as the effort of a driver to avoid interrupting other's original plan, i.e., minimizing the difference of other car's impacted behavior distribution in presence of ego car (\cref{fig:reward}(c)) and other car's original behavior distribution in absence of the ego car (\cref{fig:reward}(d)). 

Intuitively, the ego car's action could change the other car's environment and influence the other car's decision. Thus, as in \cref{fig:reward}(c), the other car's behavior distribution in the presence of the ego car is exactly given as in \eqref{eq: prediction under ego car}.

Similarly, we can write out the other car's behavior distribution in the absence of the ego car as follows:
\begin{equation}\label{eq:prediction_absence}
	P({\bf{u}}_O|{x}_O^0)=\frac{e^{R_O(x_O^0, {\bf{u}}_O)}}{\sum_{{\bf{u}}_O \in {\bf{U}}_O}e^{R_O(x_O^0, {\bf{u}}_O)}},
\end{equation}
where the utility function $R_O(x_O^0, {\bf{u}}_O)$ captures the other car's utilities (efficiency and comfort) in the absence of the ego car, as demonstrated in \cref{fig:reward}(d).

With \eqref{eq: prediction under ego car} and \eqref{eq:prediction_absence}, we are able to model the courtesy as the difference between the two distributions. We adopt the Kullback-Leibler divergence \cite{kullback1997information} as a distance metric. Therefore, we have
the courtesy reward of ego car's action ${\bf{u}}_E$ defined as
\begin{align}
\label{eq:courtesy_rw}
	R_{E_{courtesy}}(\textbf{x}^0, {\bf{u}}_E) &= e^{-D_{KL}\left[P({\bf{u}}_O|x^0_O)||P({\bf{u}}_O|{\bf{x}}^0, {\bf{u}}_E)\right]}\\ 
	&= e^{-\sum_{{\bf{u}}_O \in {\bf{U}}_O}P({\bf{u}}_O|x^0_O)\log{\frac{P({\bf{u}}_O|x^0_O)}{P({\bf{u}}_O|{\bf{x}}^0, {\bf{u}}_E)}}}.\nonumber
\end{align}
\subsection{Confidence Reward}
\label{subsec:Confidence Reward}
The confidence reward is defined to capture an agent's preference for certainty rather than randomness in others' behavior. Agents prefer to gain confidence by constraining other's behavior to a concentrated distribution.

Several measures can be used to quantify confidence, including maximum model\cite{kepecs2012computational} (maximum probability), entropy model\cite{kullback1997information} (negative entropy of the distribution), and difference model\cite{li2020confidence} (the difference between the two highest probabilities). Though the three models are consistent for two-option tasks, it is proved that the difference model can better measure confidence in multiple-option decision-making tasks \cite{li2020confidence}. Interactive driving with a discrete behavior space is exactly a multiple-option decision-making task.
Hence, we adopt the difference model and define the confidence under ego car's action $\mathbf{u}_E$ as
\begin{equation}
\label{eq:confidence}
 Conf({\bf{x}}^0, {\bf{u}}_E)=P({\bf{u}}^1_O|{\bf{x}}^0, {\bf{u}}_E) - P({\bf{u}}^2_O|{\bf{x}}^0, {\bf{u}}_E)
\end{equation}
where ${\bf{u}}^1_O$ and ${\bf{u}}^2_O$ are the two responsive actions from the other agent which, respectively, takes the highest and second-highest probabilities.

Therefore, as in \cref{fig:reward}(e), the confidence reward of the ego car's action ${\bf{u}}_E$ can be expressed as:
\begin{equation}
	R_{E_{confidence}}({\bf{x}}^0, {\bf{u}}_E) = e^{Conf({\bf{x}}^0, {\bf{u}}_E)}.\label{eq:confidence_rw}
\end{equation}

%% file: 4estimate.tex
\section{Online Estimate on Reward Parameter}
\label{sec:online estimate}
In \cref{Sec:reward design}, we have designed a socially compatible reward to capture how humans drive. However, humans are diverse in terms of their preferences over the three reward terms. Hence, we should infer each individual's preference by estimating the reward parameter from historical observations.

The key idea of the online identification is based on the principle of maximum entropy, i.e., the probability of one candidate reward parameter $\lambda$ is proportional to the probability of the observed trajectories under this reward parameter. Hence, if we look $r$ steps to the past from current time step $k$, then the posterior probability of $\lambda$ satisfies
\begin{equation}\label{eq:bayesian}
p({\bf{\lambda}}|\hat{\bf{x}}^{k-r:k}) \propto p({\hat\bf{x}}^{k-r:k}|{\bf{\lambda}}) \times p({\bf{\lambda}})
\end{equation}
where $p({\bf{\lambda}})$ is the prior probability of the candidate reward parameter. The probability of the observed trajectories $\hat{\bf{x}}^{k-r:k}$ under the candidate reward parameter is proportional to its normalized collected reward:
\begin{align}
\label{Eq:update probability}
p(\hat{\bf{x}}^{k-r:k}|\lambda)\approx \frac{e^{R({\hat{{\bf{u}}}^{k-r:k}}, \lambda)}}{\sum_{{{\bf{u}}\in \textbf{U}}}{e^{R({{{\bf{u}}}^{k-r:k}}, \lambda)}}}
\end{align}
where $\textbf{U}$ is the set of all possible actions, and $\hat{{\bf{u}}}$ is the action generating trajectory closest (determined by mean squared error) to the observed historical trajectory.

Based on \eqref{eq:bayesian} and \eqref{Eq:update probability}, we develop a Bayesian inference algorithm to estimate the reward parameter online. The algorithm is outlined in Algorithm \ref{alg:policy inference}.
We first sample N reward parameters ${\bf{\lambda}_i}$ in Line~\ref{code:initial} and initialize their weights $\omega_i(\lambda_i)$ randomly or based on the prior knowledge $\Theta{_0}(\widehat{\omega})$. After updating the weights in Line~\ref{code:update} and normalizing the weights in Line~\ref{code:normalize}, Line~\ref{code:estimate} estimates the reward parameter $\hat{\lambda}^{k-r}$ as the mean of the posterior distribution.
\begin{algorithm}[t]
	\caption{Policy Bayesian Inference}
	\label{alg:policy inference}
	\begin{algorithmic}[1]
		\Require
		N reward parameter samples \textbf{${\lambda}{_i}$},  corresponding weights ${\omega}_i{^{k-r-1}}$, and observations $\hat{\textbf{x}}{^{k-r:k}}$
		\Ensure
		Estimated reward parameter \textbf{$\check{{\lambda}}{^{k-r}}$}, updated weights for each reward parameter sample ${\omega}_i{^{k-r}}$,
		\If{k-r-1=0}
    		\State Initialize N reward parameter samples ${\lambda}{_i}$ and their corresponding weights $\omega_i^0 \sim \Theta{_0}(\widehat{\omega})$ \label{code:initial}
		\EndIf
		\For{all N reward parameter samples}
    		\State Update ${\omega}{^{k-r}_{i}} \leftarrow {\omega}{^{k-r-1}_{i}} \times
    		p(\widehat{\textbf{x}}{^{k-r:k}\vert\lambda_i})$, Eq~\ref{Eq:update probability}
    		\label{code:update}
		\EndFor
		\State Normalize ${{\omega}}{^{k-r} \leftarrow {{\omega}}{^{k-r}/\sum\nolimits^N_{i=1}}{\omega}{^{k-r}_{i}}}$
		\label{code:normalize}
		\State Compute ${\check{\lambda}}^{k-r} \leftarrow\sum\nolimits^N_{i=1}\lambda_{i}{\omega}{^{k-r}_{i}}$
		\label{code:estimate}
	\end{algorithmic}
\end{algorithm}

%% file: 5experiments.tex
\section{Experimental Results}
	\begin{figure*}[!ht]
    	\centering
        \includegraphics[width=0.9\textwidth]{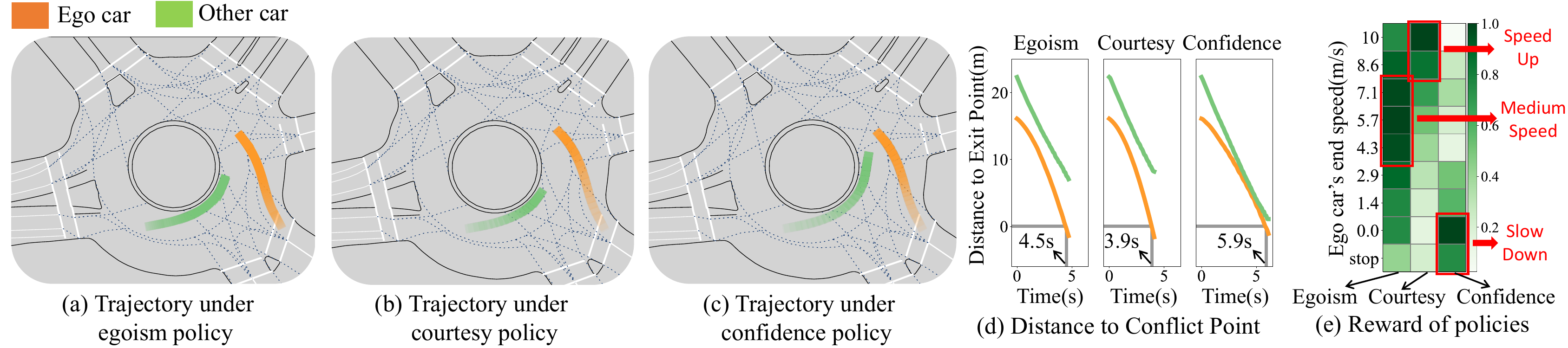}
        \caption{The interaction when the ego car initially preceded. (a)(b)(c) show the interaction trajectories when the ego car takes different policies and the other car runs on the egoism policy. (d) shows the two cars' distances to the conflict point. (e) shows the rewards of three policies at the first planning step. Darker color means a higher reward. The egoistic car run at a medium speed, distance, and interaction period. The courteous car would speed up to give more room, which generated the largest relative distance and shortest interaction period. The confident car would slow down to block the other car, resulting in a smaller relative distance and a longer interaction period.}
        \label{fig:Analysis ego precedes}
        \label{online estimate}
        \vspace{-0.5em}
    \end{figure*}
\label{sec:result}
   Two sets of experiments were conducted. First, we analyzed the influences of the three reward terms on interactive behaviors with varying relative relationships between the two vehicles. Second, we deployed the online inference algorithm on real human driving data, and compared the re-generated trajectories via the proposed UAPP framework against the ground truth. The statistical results of humans' driving behaviors in different scenarios, traffic rules, and countries were collected and discussed.

    \begin{figure*}[!ht]
    	\centering
        \includegraphics[width=0.9\textwidth]{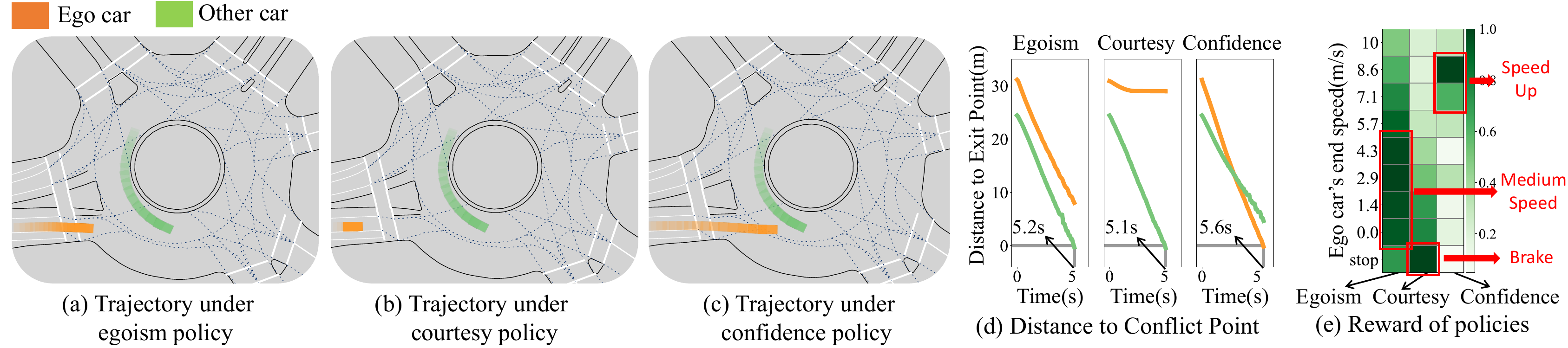}
        \caption{The interaction when the other car initially preceded. Compared to an egoistic car, a courteous ego car would brake to give way to the other car. Contrarily, an overly confident car would speed up to chase the other car, which resulted in the smallest relative distance and longest interaction period.}
    	\label{fig:Analysis other precedes}
    	\vspace{-1em}
    \end{figure*}

    \textbf{Experiment Setting}: We selected three roundabout scenarios from the INTERACTION dataset\cite{zhan2019interaction}: USA Roundabout FT, USA Roundabout SR, and DEU Roundabout OF. Totally we chose 219 pairs of interactive trajectories (168, 25, and 26 from each scenario). Each pair contains two cars, with one car trying to merge into a roundabout while the other car driving in the roundabout. The two cars drove towards a conflict point (the intersection point on the two cars' reference paths). The experiments were conducted in Robot Operation System (ROS) on a computer with a 3.60GHz Intel Core i7 processor of 16GM RAM. To focus on the social-compatible planning, throughout the simulation, we assume that both cars have perfect conditions on map information, perception, localization, and control.
    \subsection{Influences of Courtesy and Confidence to Behaviors}
    \label{SubSec:Analysis of Policy}
   In this experiment, we simulated the interaction between two vehicles in the USA Roundabout FT scenario. The ego car was set as the car merging into the roundabout, and the other car was in the roundabout. Both cars were AVs, while the ego car was the leader and the other car was a follower. The ego car drove actively via three different policies: 1) an egoism policy that only cares about its egoism reward, 2) a courtesy policy that cares only about courtesy, i.e., its potential influence on others, and 3) a confidence policy that generates behaviors purely for higher confidence. The other car was assumed to precisely knew the ego car's actions and was simply reacting by an egoism policy as a pure follower.
    \subsubsection{Quantitative Results}
    In the USA Roundabout FT scenario, we generated interactive trajectories by setting the initial states of both cars as given by the ground truth. Two metrics are calculated: averaged relative distance (ARE) between the two vehicles during the interaction, and the averaged interaction time (AIT, the time span from start to the time instant when one car passed the conflict point). As shown in Tab.~I, courteous cars generated the safest and most efficient interaction. Though it appears counter-intuitive, it proved that via cooperation, more efficient interactions would happen. Confidence-driven cars could result in dangerous situations and were the most time-consuming because many unnecessary acceleration-deceleration were generated to assure safety. Egoistic cars performed between confident-driven cars and courteous cars.
    \begin{figure*}[!ht]
    	\centering
    	\includegraphics[width=0.9\textwidth]{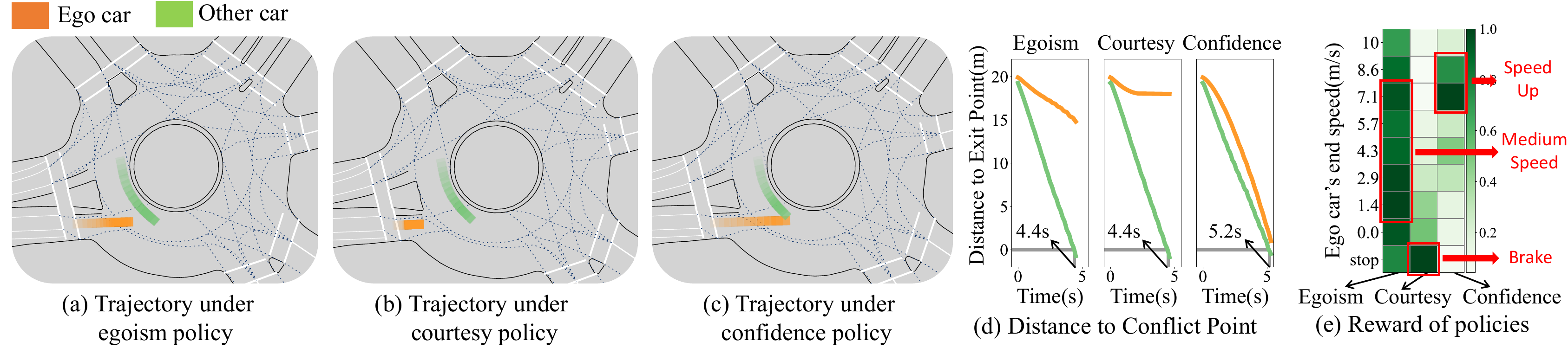}
    	\caption{The interaction when the two cars were initially equally positioned. Similarly, a courteous car would brake to make space for the other car, and an overly confident car speeded up to constrain the other car's behavior, resulting in a smaller distance and longer interaction period.}
    	\label{fig:Analysis equal position}
    	\vspace{-0.5em}
    \end{figure*}
    \begin{figure*}[!ht]
    	\centering
    	\includegraphics[width=0.9\textwidth]{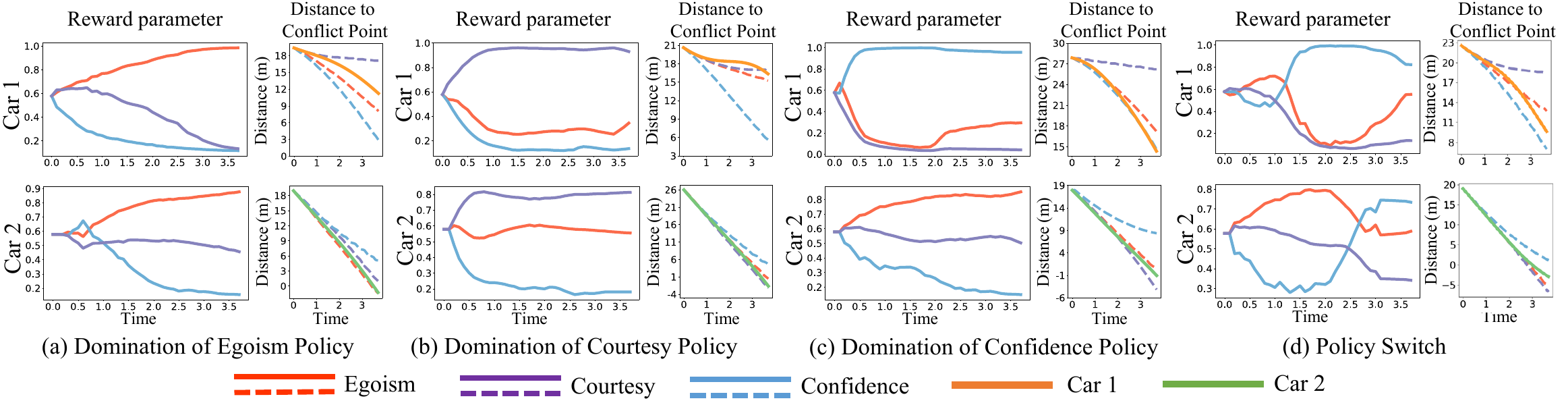}
    	\caption{Exemplary results of estimating reward parameter of real driving data, where the human driver is dominated by one policy in (a)(b)(c) and the policy switch happens in (d). For each example, we show how the reward parameter evolved and two cars' distances to conflict point.}
    	\label{fig:Estimate Examples}
		\vspace{-0.8em}
    \end{figure*}

	\begin{table}[!t]
		\label{tab:influence_table}
		\centering
		\begin{tabular}{cccccc} 
			\toprule 
			& Egoism & Courtesy & Confidence\\
			\midrule 
			ARE (m) & 8.74 & 10.26 & 5.74\\
			AIT (s)     & 4.63 & 4.38  & 5.24\\
			\bottomrule 
		\end{tabular} 
		\caption{The quantitative results for the influence of different policies. Courtesy is the safest and most efficient. Confidence is the most dangerous and time-consuming. Egoism is between the other two policies.} 
		\vspace{-2em}
	\end{table}

    \subsubsection{Case Analysis}
    With Fig.~\ref{fig:Analysis ego precedes}-\ref{fig:Analysis equal position}, we provide a detailed analysis of such influences in three representative situations where the two vehicles started the interaction with different initial positions: Case I - the ego car initially preceded (i.e., closer to the conflict point) in \cref{fig:Analysis ego precedes}, Case II - the other car initially preceded (\cref{fig:Analysis other precedes}), and Case III - the two cars initially equally positioned (\cref{fig:Analysis equal position}). We can see from (e) that in all three cases, the egoism policy preferred medium speeds, while the influence of the courtesy policy and the confidence policy varied. In Case I where the ego car preceded, the courtesy policy encouraged the ego car to speed up to leave more space for the other car, while the confidence policy preferred slower speeds to reduce the variance of the other car's possible actions. However, in Case II and III when the other car preceded or the two cars were equally positioned, the courtesy policy enabled the ego car to give way to the other car, and the confidence policy drove the ego car to chase for the other car so that it was less likely to decelerate.

\subsection{Online Estimation of Reward Parameters}
\label{sec:behavior generation}
The second study we conducted was to infer human drivers' policies in the interactive human data of the USA Roundabout FT scenario. By implementing the Bayesian inference algorithm in Algorithm~\ref{alg:policy inference}, we can online estimate the weights on the three rewards (egoism, courtesy, and confidence) for each individual and re-generate human-like behaviors via the UAPP framework.
\begin{figure}[!t]
    \centering
    \includegraphics[width=0.48\textwidth]{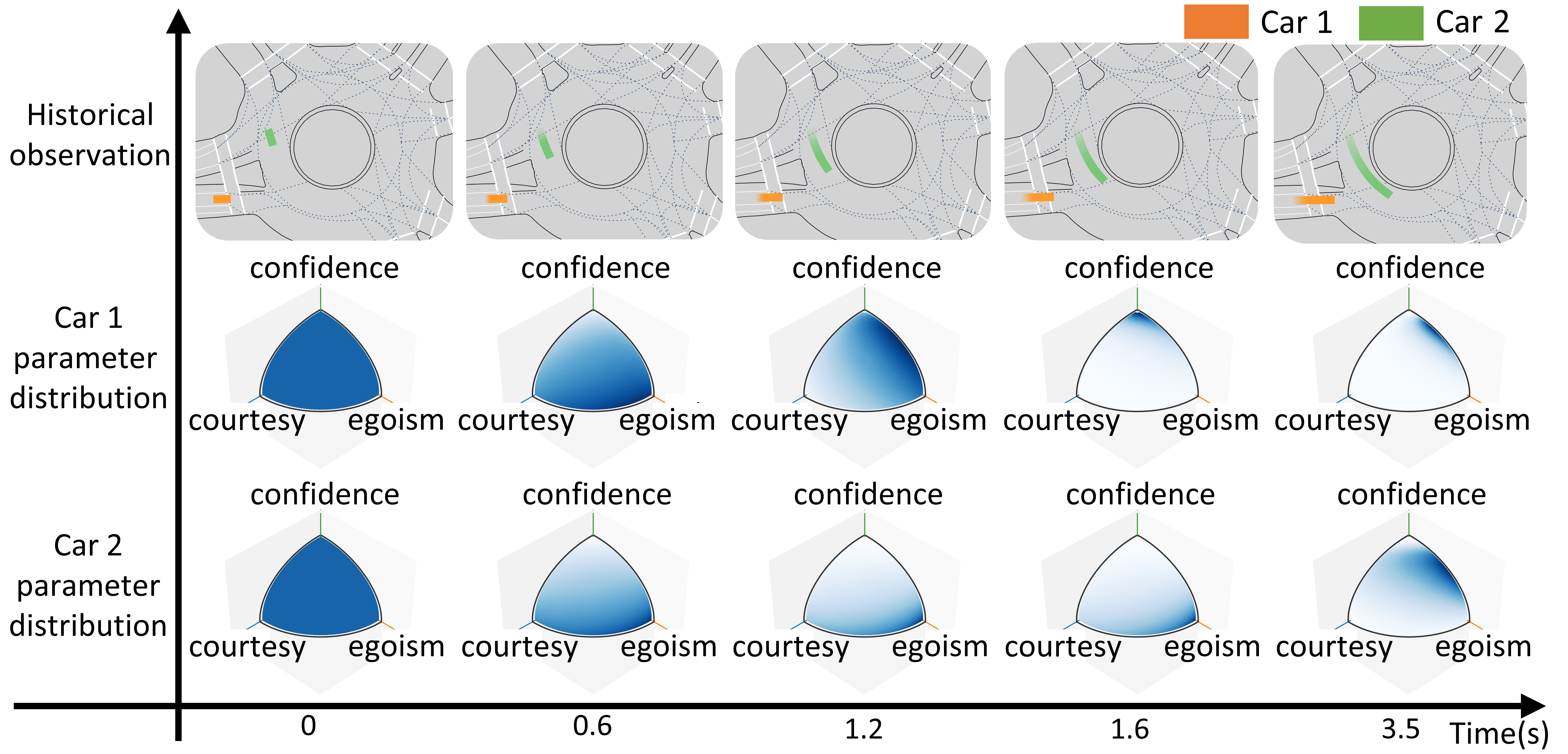}
    \caption{Visualizations of the process on estimating reward parameters from historical observations. The parameters converge with observations.}
    \label{fig:Estimate Process}
    \vspace{-1em}
\end{figure}
\Cref{fig:Estimate Process} shows the estimate process in one example. We run the algorithm on each car separately. We can see that with more and more observations, the samples representing the reward weights converge. 

Several exemplary results are shown in \cref{fig:Estimate Examples}. In each subfigure, we show, respectively, the estimates of the two cars' reward parameters, their distances to the conflict point. Car 1 and car 2 denote, respectively, the drivers outside and inside the roundabout.

\textbf{Egoism Policy Dominates.} In \cref{fig:Estimate Examples}(a), both cars took mediate speeds. So both cars are inferred as being egoistic.

\textbf{Courtesy Policy Dominates.} In \cref{fig:Estimate Examples}(b), car 1, which entered the roundabout, almost braked thoroughly with a strong yielding intention, while car 2 took a high speed trying to finish the interaction. Both cars were interpreted as being courteous enough to not change others' original plans.

\textbf{Confidence Policy Dominates.} In \cref{fig:Estimate Examples}(c), though running after car 2, car 1 still took a high speed, which was inferred as being confident that car 2 would constantly keep a high speed ahead. Car 2, on the other hand, was identified as running an egoistic policy for its mediate speed.

\textbf{Policy Switch Exists.} We also observed that agents switch policies. As in \cref{fig:Estimate Examples}(d), initially, car 1 and car 2 both took mediate speeds with an egoism policy. But later car 1 speeded up, being confident that car 2 would keep a high speed to leave the roundabout. Meanwhile, car 2 decelerated, confident that car 1 would not chase it. Thus the confidence policy outperformed the egoism policy, and a policy switch happened.

     \begin{table*}[h]
	\begin{subtable}[h]{\textwidth}
		\centering
		\begin{tabular}{ccccccccc} 
			\toprule 
Horizon(s)  & Egoism & Courtesy & Confidence & *Egoism & *Courtesy & *Confidence & *Uniform & *DOP\\ 
			\midrule 
            0.3 & 1 (0.01407) & 0.2\%$\uparrow$ & 0.4\%$\uparrow$ & \textbf{8.0\%$\downarrow$} & \textbf{6.7\%$\downarrow$} & \textbf{7.1\%$\downarrow$} & \textbf{5.6\%$\downarrow$} & \textbf{7.0\%$\downarrow$}\\
            0.5 & 1 (0.03366) & 0.4\%$\uparrow$ &	0.3\%$\uparrow$ &	\textbf{14.5\%$\downarrow$} & \textbf{12.2\%$\downarrow$} & \textbf{13.8\%$\downarrow$} & \textbf{11.3\%$\downarrow$} &	\textbf{13.2\%$\downarrow$}\\
            1   & 1 (0.11204) & 4.2\%$\uparrow$ &	3.6\%$\uparrow$ &	\textbf{18.8\%$\downarrow$} & \textbf{16.3\%$\downarrow$} & \textbf{18.9\%$\downarrow$} & \textbf{16.1\%$\downarrow$} &	\textbf{18.9\%$\downarrow$}\\
			\bottomrule 

		\end{tabular} 
		\caption{Outside roundabout} 
	\end{subtable}
	\begin{subtable}[h]{\textwidth}
		\centering
        \begin{tabular}{ccccccccc} 
		\toprule 

Horizon(s)  & Egoism & Courtesy & Confidence & *Egoism & *Courtesy & *Confidence & *Uniform & *DOP\\ 
            \midrule 
            0.3 & 1 (0.01964) & 0.0\% & 5.8\%$\uparrow$ & \textbf{0.2\%$\downarrow$} & 0.7\%$\uparrow$ & 0.8\% $\uparrow$ & 0.1\%$\uparrow$ & \textbf{0.9\%$\downarrow$}\\
            0.5 & 1 (0.03031) & 2.9\%$\uparrow$ &	24.2\%$\uparrow$ &	\textbf{0.5\%$\downarrow$} & 4.1\%$\uparrow$ & 4.4\%$\uparrow$ & 2.3\%$\uparrow$ & \textbf{1.7\%$\downarrow$}\\
            1   & 1 (0.08001) & 10.7\%$\uparrow$ & 54.0\%$\uparrow$ &	0.1\%$\uparrow$ &	10.6\%$\uparrow$ & 11.2\%$\uparrow$ &	7.6\%$\uparrow$ &	\textbf{3.4\%$\downarrow$}\\
		\bottomrule 
		\end{tabular}
		\caption{Inside roundabout} 
	\end{subtable}
	\begin{subtable}[h]{\textwidth}
		\centering
		\begin{tabular}{ccccccccc} 
			\toprule 
		Horizon(s)  & Egoism & Courtesy & Confidence & *Egoism & *Courtesy & *Confidence & *Uniform & *DOP\\ 
\midrule 
0.3 & 1 (0.01613) & \textbf{37.9\%$\downarrow$} & 5.8\%$\uparrow$ & \textbf{31.7\%$\downarrow$} & \textbf{32.9\%$\downarrow$} &	\textbf{4.6\%$\downarrow$} & \textbf{38.0\%$\downarrow$} & \textbf{31.2\%$\downarrow$}\\
0.5 & 1 (0.04678) & \textbf{55.2\%$\downarrow$} & 13.9\%$\uparrow$ & \textbf{44.1\%$\downarrow$} & \textbf{48.2\%$\downarrow$} &	\textbf{42.4\%$\downarrow$} & \textbf{42.0\%$\downarrow$} & \textbf{46.6\%$\downarrow$}\\
1   & 1 (0.17619) & \textbf{68.7\%$\downarrow$} & 20.2\%$\uparrow$ & \textbf{54.7\%$\downarrow$} & \textbf{61.1\%$\downarrow$} &	\textbf{53.8\%$\downarrow$} & \textbf{54.4\%$\downarrow$} & \textbf{60.1\%$\downarrow$}\\
			\bottomrule 
		\end{tabular}
		\caption{Cases where Courtesy Policy Dominates} 
	\end{subtable}
	\begin{subtable}[h]{\textwidth}
    	\centering
    	\begin{tabular}{ccccccccc} 
    		\toprule 
    Horizon(s)  & Egoism & Courtesy & Confidence & *Egoism & *Courtesy & *Confidence & *Uniform & *DOP\\ 
    		\midrule 
            0.3 & 1 (0.01314) & 6.4\%$\uparrow$ & \textbf{28.2\%$\downarrow$} & \textbf{26.6\%$\downarrow$} & \textbf{24.8\% $\downarrow$} & \textbf{26.9\% $\downarrow$} &	\textbf{24.7\%$\downarrow$} & \textbf{25.5\%$\downarrow$}\\
            0.5 & 1 (0.03923) & 13.0\%$\uparrow$ & \textbf{47.4\%$\downarrow$} & \textbf{41.6\%$\downarrow$} & \textbf{39.5\% $\downarrow$} & \textbf{44.7\% $\downarrow$} &	\textbf{41.0\%$\downarrow$} & \textbf{40.9\%$\downarrow$} \\
            1   & 1 (0.14461) & 19.8\%$\uparrow$ & \textbf{60.8\%$\downarrow$} & \textbf{52.0\%$\downarrow$} & \textbf{51.0\% $\downarrow$} & \textbf{57.5\% $\downarrow$} &	\textbf{53.6\%$\downarrow$} &	\textbf{52.9\%$\downarrow$} \\
    		\bottomrule 
    	\end{tabular}
    	\caption{Cases where Confidence Policy Dominates} 
    \end{subtable}
	\caption{The MSE between the re-generated behaviors and the ground-truth under different policies. Setting the error of egoism policy as the baseline, we show how much percent the errors of other policies increase or decrease by. Arrows indicate whether the error increases or decreases. * denotes the behaviors via the online estimated parameter with different initialization. *DOP denotes the initialization based on the dominance of policy. Compared to the traditional fixed-egoism-policy method, our method with DOP initialization improved the performance by 19\% and 3\% for drivers outside and inside the roundabout, and by 60\% and 53\% for courteous and confident drivers.}
    \label{tab:all error}
    \vspace{-2em}
\end{table*}

\subsection{Statistical Results of Humans' Driver behaviors}
\label{sec:Policy Statistical Results}
To investigate how humans' driving behaviors differ according to traffic rules and cultures, we run the inference algorithm on three scenarios sampled from three different locations in the INTERACTION dataset, i.e., the USA Roundabout FT, the USA Roundabout SR, and the DEU Roundabout OF.
\subsubsection{Metrics} We calculated two metrics to draw a general understanding of human drivers' policies.
 
\textbf{Policy switching frequency (PSF)} counts how many times a human driver switches his/her policies during one interaction. \Cref{fig:Statistical}(a) shows that, both drivers inside and outside the roundabout mostly switched policies no more than 3 times. Such results can serve as prior knowledge that we should not design behavior with frequent policy switch and should not expect humans might switch policy too often during one interaction.
 
 \textbf{Dominance of policy (DOP)} quantifies how long each policy is serving as a dominant policy. 
 \subsubsection{Influence of traffic rules and right-of-ways} Traffic rules and right-of-ways for road users could significantly affect humans' driving behaviors. In the USA Roundabout FT, all the entrances set a stop sign, requiring a full stop before entering the roundabout. Hence, we can see that in \cref{fig:Statistical}(b), cars trying to merge-in are highly courteous (49\%) and much less egoistic (27\%) and confidence (23\%) because they do not have the right-of-way. Meanwhile, drivers inside the roundabout were mostly driving with an egoism policy (46\%) since they hold the right-of-way. However, they were also showing a great percentage of courtesy (43\%) because they tended to leave the roundabout faster to let drivers outside the roundabout merge in sooner. They were less confident (9\%).
 
 On the contrary, in the USA Roundabout SR where yield signs are set at each entrance, outside cars are driving with a much lower courtesy weight (31\%), as shown in \cref{fig:Statistical}(c). Cars inside the roundabout behave similarly in the two scenarios because the two roundabouts set the same traffic rules for them. 
 
  \subsubsection{Influence of cultures} 
\begin{figure}[!ht]
    \centering
    \includegraphics[width=0.47\textwidth]{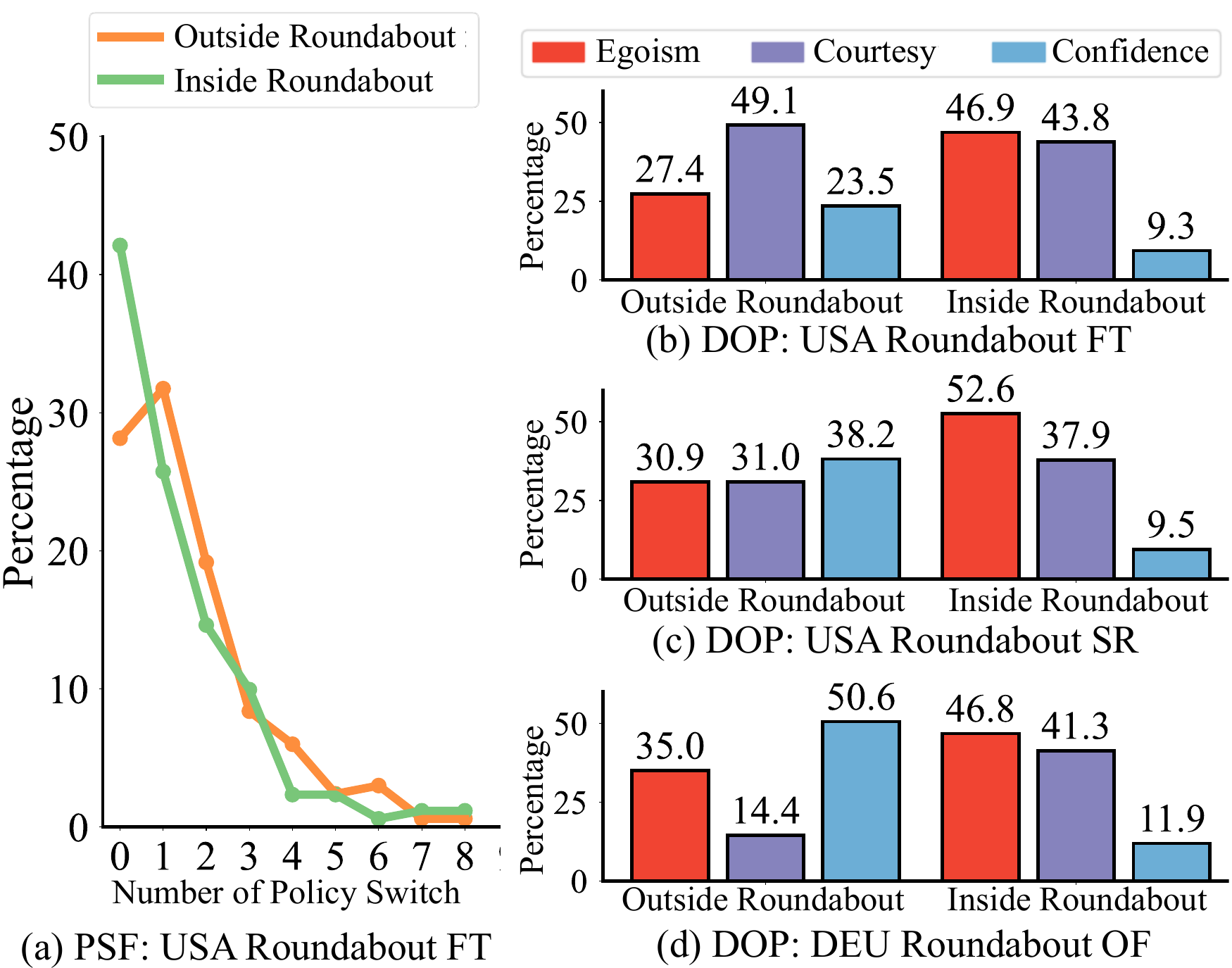}
    \caption{(a) shows the policy switch frequency(PSF) on real data in USA Roundabout FT. Most drivers switched no more than 3 times. (b)(c)(d) show the dominance of policy(DOP) of real data in three scenarios. Driving preference varies with right-of-ways, traffic rules, and cultures.}
    \label{fig:Statistical}
     \vspace{-2em}
\end{figure}

Human behavior is also significantly impacted by cultural factors \cite{hofstede2001culture}. To investigate such impacts, we inferred humans' driving preferences on two scenarios located in different countries. The USA Roundabout SR (in the US) and the DEU Roundabout OF (in Germany) scenarios were selected in the INTERACTION dataset. Both roundabouts have the same traffic rules (yield sign at each entrance). In \cref{fig:Statistical}(c)(d), we can see that when outside drivers are entering the roundabout with a yield sign, drivers in the USA are driving on the three policies almost equally(egoism 30.9\%, courtesy 31.0\%, confidence 38.2\%), while about half of the drivers in Germany are driving with high confidence (50.6\%) and much less care about courtesy (14.4\%). The distributions of driving styles inside the roundabout are similar in the two countries.
  
  These results indicate a practical application of our algorithm on the traffic rule scheduling and culture study. More applications such as responsibility assessment in traffic accidents, usage-based vehicle insurance could also be expected.

\subsection{Human-like Behavior Generation}
With the estimated reward parameter, we are also able to re-generate the interactive behaviors via the proposed integrated prediction and planning framework. To evaluate the human-likeness, we compare the difference between the re-generated trajectories and the ground-truth. To show the influence of different policies, we re-generated four sets of behaviors, i.e., 1) under egoism policy, 2) under courtesy policy, 3) under confidence policy, and 4) under the mixed policy with online estimated reward weights.

Several initialization strategies were explored, including initializing with a higher egoism, a higher courtesy, a higher confidence policy, uniform initialization, and the statistical results given by the dominance of policy in \cref{sec:Policy Statistical Results}. We calculated the mean squared error (MSE) between the ground-truth and predicted trajectories.

From \cref{tab:all error}(a), we can see that for cars outside the roundabout where non-egoistic behaviors were frequent, all the re-generated behaviors under the mixed policy with online estimated reward weights were more human-like, achieving the smallest MSE compared to the ground truth. From \cref{tab:all error}(b), we can tell that for cars inside the roundabout who were mostly egoistic,  different initialization had a significant impact on the performance of human-likeness. Among the five initialization settings, the one given by the statistical results (DOP in \cref{sec:Policy Statistical Results}) generated the best performance.

Furthermore, we selected cases where non-egoistic policies, namely the courtesy policy, and the confidence policy, strongly dominated (weight of the policy $\lambda_{policy}$ exceeds 0.9 for at least half of the interaction time). As shown in Table.~\ref{tab:all error}(c)(d), for all different initializations, the behavior generation algorithm with online-estimated reward weights performed much better than typically using an egoism policy. The initialization with statistical results (DOP in \cref{sec:Policy Statistical Results}) reduced the MSE by 60\% and 53\% respectively in cases dominated by courtesy and confidence.